\documentclass[a4paper,fleqn]{cas-dc}

\usepackage[numbers]{natbib}
\usepackage{adjustbox}
\usepackage[ruled]{algorithm2e}

\def\tsc#1{\csdef{#1}{\textsc{\lowercase{#1}}\xspace}}
\tsc{WGM}
\tsc{QE}

\begin{document}
\let\WriteBookmarks\relax
\def\floatpagepagefraction{1}
\def\textpagefraction{.001}

\shorttitle{Cross-Modal Concept Learning and Inference for Vision-Language Models}    

\shortauthors{Y. Zhang, C. Zhang, Y. Tang, and Z. He}  

\title [mode = title]{Cross-Modal Concept Learning and Inference for Vision-Language Models}  



%

\author[1,2]{Yi Zhang}[orcid=0000-0002-5831-0170]


\affiliation[1]{
organization={Harbin Institute of Technology},
city = {Harbin},
citysep={}, 
postcode={150001}, 
country={China}}
\affiliation[2]{
organization={Southern University of Science and Technology},
city = {Shenzhen},
citysep={}, 
postcode={518055}, 
country={China}}
\affiliation[3]{organization={Pengcheng Laboratory},
    city={Shenzhen},
    citysep={}, 
    postcode={518000},
    country={China}}
            
\author[2]{Ce Zhang}[orcid=0000-0001-6789-0130]

\author[2]{Yushun Tang}[orcid=0000-0002-8350-7637]

\author[2,3]{Zhihai He}[orcid=0000-0002-2647-8286]
\cormark[1]
\ead{hezh@sustech.edu.cn}


\cortext[1]{Corresponding author.}



\begin{abstract}
Large-scale pre-trained Vision-Language Models (VLMs), such as CLIP,  establish the correlation between texts and images, achieving remarkable success on various downstream tasks with fine-tuning. In existing fine-tuning methods,  
the class-specific text description is matched against the whole image. We recognize that this whole image matching is not effective since images from the same class often contain a set of different semantic objects, and an object further consists of a set of semantic parts or concepts. Individual semantic parts or concepts may appear in image samples from different classes. To address this issue, in this paper, we develop a new method called cross-model concept learning and inference (CCLI). Using the powerful text-image correlation capability of CLIP, our method automatically learns a large set of distinctive visual concepts from images using a set of semantic text concepts. Based on these visual concepts, we construct a discriminative representation of images and learn a concept inference network to perform downstream image classification tasks, such as few-shot learning and domain generalization. 
Extensive experimental results demonstrate that our CCLI method is able to improve the performance upon the current state-of-the-art methods by large margins, for example, 
 by up to 8.0\% improvement on few-shot learning and by up to 1.3\% for domain generalization. 
\end{abstract}




\maketitle


\section{Introduction}
\label{sec:intro}
Recently, large-scale pre-trained Vision-Language Models (VLMs) emerges as an important research topic, which has achieved remarkable success on various downstream tasks \cite{radford2021learning,zhou2022learning,gao2021clip}. Compared to traditional methods, those pre-trained VLMs encode and map texts and images into a unified space, resulting in better transfer capabilities \cite{gao2021clip,yu2022coca}.
Pre-trained VLMs such as CLIP \cite{radford2021learning}  establish the powerful connection between texts and images. 
It should be noted that since these pre-trained VLMs are of massive sizes and computationally impractical to re-train. Thus, it remains a challenging task to adapt the well-learned knowledge of VLMs to downstream tasks.

To address this issue, a number of approaches have been developed to efficiently adapt such models with very limited supervision. Those approaches can be classified into two categories, namely, prompt tuning methods \cite{zhou2022learning,zhou2022conditional,shu2022tpt,lu2022prompt,wang2022learning,jia2022vpt} and adapter-based methods \cite{gao2021clip,zhang2022tip,svladapterbmvc2022}. The prompt tuning methods, such as CoOp~\cite{zhou2022learning} and CoCoOp~\cite{zhou2022conditional}, focus on designing delicate prompts and introducing learnable context to distill the task-relevant information from the rich knowledge encoded in CLIP. In contrast, adapter-based methods, such as CLIP-Adapter \cite{gao2021clip} and Tip-Adapter \cite{zhang2022tip}, fine-tune the representations generated by CLIP's encoders to better represent images and texts.

\begin{figure}[!t]
\centering
\includegraphics[width=\linewidth]{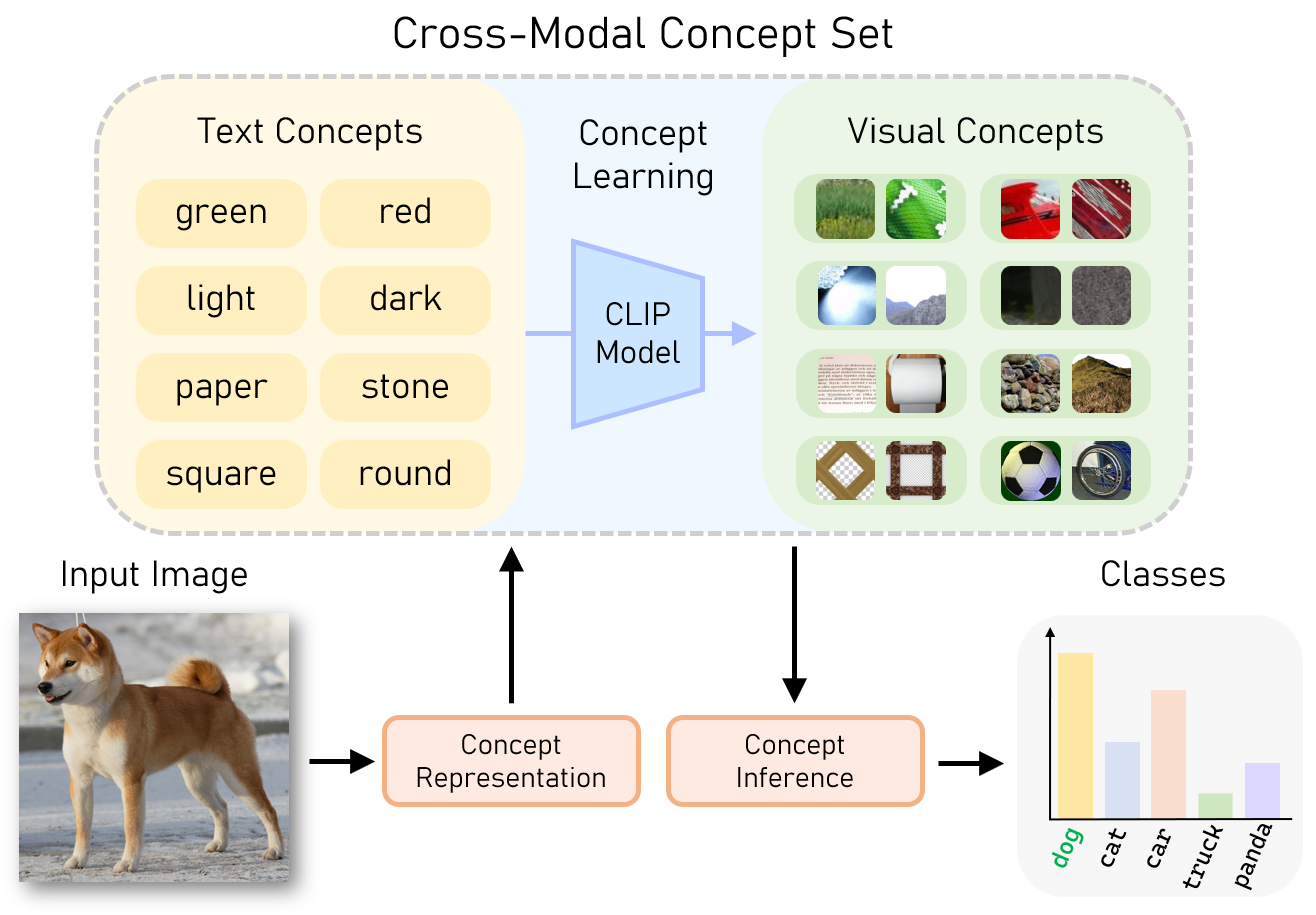}
\caption{\textbf{An illustration of our major idea}. Empowered by CLIP, we build a cross-modal concept set to enable concept-level representations for input images to obtain more accurate classification results.}
\label{fig:codebook}
\vspace{-15pt}
\end{figure}

We observe that in the current fine-turning methods for CLIP, the class-specific text is matched against the whole image. We recognize that this matching method is not effective because: (1) images from the same class often contain a set of different semantic objects which correspond to different text descriptions. (2) An object further consists of a set of different semantic parts which also have different text descriptions. (3) On the other hand, individual semantic objects and parts, known as concepts, may appear in image samples from different classes. For example, the \texttt{Cat} and \texttt{Car} images  may both contain the tree object. The \texttt{Car}  and \texttt{Truck} images may both contain the semantic part of wheels or have the same color concepts. 
This mixture of visual concepts in a natural image 
will cause major problems when we attempt to match the class-specific text description against the whole image.
This problem, if not efficiently addressed, will hinder our capabilities in general image-text understanding and downstream tasks.  

To address this important issue, in this paper, we establish and learn a semantic concept-level representation and inference of the image-text pairs. This method, called cross-modal concept learning and inference (CCLI), provides a new approach to explore the potential of CLIP for correlating text and images. 
Specifically, as illustrated in Figure \ref{fig:codebook}, based on the powerful text-image correlation capability of CLIP, our method automatically learns a large set of distinctive visual concepts from images using a set of pre-defined semantic text concepts. Based on these visual concepts, we construct a discriminative representation of images and learn a concept inference network to perform downstream image classification tasks. 
We observe that this concept-level representation and inference is able to provide better consistency between vision and language modalities, resulting in a much-improved generalization capability. 
The proposed CCLI  method is successfully applied to few-shot image classification and domain generalization tasks. It achieves significantly improved performance, outperforming the current state-of-the-art methods by large margins, for example, by up to 8.0\% improvement on few-shot learning and by up to 1.3\% for domain generalization.

\section{Related Work}
\label{sec:related}
In this section, we review related works on large-scale pre-trained VLMs, fine-tuning for VLMs, few-shot image classification, generalization under distribution shift, and visual concept learning. 

\textbf{(1) Large-scale pre-trained VLMs.}
Large-scale pre-trained VLMs have been developed to learn general visual representation under the supervision of natural languages  \cite{lei2015predicting,gomez2017self,sariyildiz2020learning,desai2021virtex,radford2021learning}. 
Recent research has explored the semantic correspondence between the linguistic and visual modalities by leveraging a large number of image-text pairs available on the internet  \cite{jia2021scaling,radford2021learning,yu2022coca}. 
For instance, CLIP \cite{radford2021learning} is obtained by contrastive learning on 400 million curated image-text pairs, while ALIGN \cite{jia2021scaling} exploits 1.8 billion noisy image-text pairs from the raw alt-text data. 
There has been several other works studying along the direction of large-scale VLMs, examples include CoCa \cite{yu2022coca}, SimVLM \cite{wang2022simvlm}, ZeroVL \cite{cui2022contrastive} Florence \cite{yuan2021florence}, BEiT \cite{wang2022image}, Flamingo \cite{alayrac2022flamingo}, GIT \cite{wang2022git}, PaLI \cite{chen2022pali} and HiCLIP \cite{geng2023hiclip}.

Researchers have demonstrated that large-scale pre-trained VLMs can be applied to a variety of cross-modal alignment, zero-shot and few-shot image recognition tasks \cite{radford2021learning,gao2021clip,zhang2021vt,zhou2022learning}, as well as other visual tasks including image retrieval \cite{lu2019vilbert,duan2022multi}, visual grounding \cite{li2021grounded,yao2021cpt}, visual question answering \cite{zhou2022unsupervised,duan2022multi,lei2021less} and image manipulation \cite{kim2022diffusionclip}. Despite the success of VLMs in many downstream applications, recent studies have also highlighted concerns regarding VLMs' ability to comprehend relation, attribution, and order \cite{yuksekgonul2023when}.

\textbf{(2) Fine-tuning for VLMs.}
Fine-tuning is crucial for VLMs to adapt to various downstream tasks. In this work, we mainly focus on the image classification task. Recent works can be categorized into prompt tuning methods and adapter-based methods. 

 \textbf{\textit{Prompt tuning methods}} are motivated by the success of prefix-tuning within the domain of natural language processing \cite{deng2022rlprompt,gao2021making,jiang2020can,liu2023pre}. As the seminal work in this field, CoOp \cite{zhou2022learning} enhances the prompt context by optimizing it through the utilization of a collection of trainable vectors.  Zhou \textit{et al.} \cite{zhou2022conditional} extends the CoOp method to address the generalization problem for unseen classes by learning to generate vectors conditioned on each image. To avoid prompt tuning from forgetting the general knowledge, ProGrad~\cite{zhu2022prompt} proposes to update the prompts whose gradients are well aligned. A number of other studies have also investigated the approach of prompt tuning for VLMs. For instance, TPT \cite{shu2022tpt} has the capability to dynamically learn adaptive prompts in real time on the fly with a single test sample. ProDA \cite{lu2022prompt} acquires low-bias prompts from a limited number of samples, effectively capturing the distribution of diverse prompts to accommodate the diverse visual representations encountered. DeFo \cite{wang2022learning} learns decomposed visual features using feature-level textual prompts. UPL \cite{huang2022unsupervised} introduces unsupervised learning into prompt learning to avoid labor-intensive prompt engineering. In addition to textual prompt tuning, VPT \cite{jia2022vpt} applies comprehensive fine-tuning techniques to large-scale transformer models specifically designed for vision tasks. UPT \cite{zang2022unified} learns a tiny neural network to jointly optimize prompts across visual and text modalities.

\textbf{\textit{Adapter-based methods}},  which are inspired by parameter-efficient finetuning methods \cite{houlsby2019parameter,zhang2020side}, directly tune the  representations generated by the CLIP's visual and text encoders. For example, CLIP-Adapter \cite{gao2021clip} proposes an additional feature adapter to boost conventional fine-tuning results. Tip-Adapter \cite{zhang2022tip} achieves enhanced outcomes by constructing a key-value cache model using low-shot samples. SAL-Adapter \cite{svladapterbmvc2022} combines the inherent strengths of vision-language pre-training and self-supervised representation learning to achieve enhanced performance.

\textbf{(3) Few-shot image classification.}
Few-shot learning has been proposed to enable generalization to new tasks with only a few supervised samples \cite{wang2020generalizing}.  Traditional few-shot learning methods leverage meta learning \cite{finn2017model}, metric learning \cite{bateni2020improved} and transfer learning \cite{qi2018low} to achieve remarkable adaptation capabilities. However, these methods typically require training from base classes in the source domain, which limits their generalization capabilities. Recent advances in pre-trained VLMs have demonstrated a promising alternative approach that does not rely on source-domain training datasets. By keeping the pre-trained weights fixed and training supplementary adaptable modules for downstream tasks, these models can achieve remarkable performance with very limited training data \cite{radford2021learning,gao2021clip,zhou2022learning,zhang2022tip,lin2023multimodality,najdenkoska2023meta}. For example, Zhang \textit{et al.} \cite{zhang2022tip} establish a key-value cache model based on the few-shot training set to serve as a weight initialization, Lin \textit{et al.} \cite{lin2023multimodality} propose a cross-modal adaptation approach to learn from few-shot instances spanning different modalities, Najdenkoska \textit{et al.} \cite{najdenkoska2023meta} defines a meta-mapper network to efficiently bridge frozen large-scale VLMs and leverage their already learned capacity.

\textbf{(4) Generalization under distribution shift.}
Distribution shift refers to the discrepancy between the distributions of training data in the source domain and test data in the target domain \cite{koh2021wilds}.
The ability to generalize to out-of-distribution (OOD) data is a natural aptitude for humans but remains a challenging task for artificial intelligence models.
To address this problem, a number of methods have been developed within the context of domain adaptation \cite{wang2018deep,liang2020we,Wang2022exploring} and test-time adaptation \cite{wang2021tent,tang2023neuro,liang2023comprehensive,kan2023self}.
In this work, we focus on domain generalization \cite{zhou2022domain,wang2022generalizing,zhou2021domain}, which aims to address the performance degradation under data distribution shifts, by training models only on the source domains that are generalizable to new unseen target domains \cite{zhou2022domain}. 
Large-scale pre-trained vision and language models, like CLIP, have showcased remarkable generalization abilities when applied to zero-shot scenarios with distribution shifts in various downstream tasks. \cite{radford2021learning}. This ability to generalize without any fine-tuning on task-specific data is a highly desirable characteristic of VLMs, and presents a promising direction for advancing machine learning methods. 

\textbf{(5) Visual concept learning.} Visual concepts/attributes have demonstrated great potential as cues for a variety of visual tasks, \textit{e.g.}, object recognition \cite{lampert2009learning,sun2013attribute,cheng2014sparse,liu2014attribute}, semantic segmentation \cite{sulistiyo2018attribute,yang2021attribute}, and zero-shot transfer \cite{russakovsky2012attribute,al2015transfer}. There are two major approaches to visual concept learning that have been explored in the existing literature. The first approach typically requires manual semantic concept labeling (such as colors, textures, and fabric) for the training images \cite{patterson2012sun,patterson2016coco,pham2021learning}. To alleviate the labeling cost, several studies \cite{amid2015multiview,nigam2019towards} propose to learn concepts under triplet supervision, where human annotators should only provide labels for similar and dissimilar objects. The second approach focuses on designing data-driven concepts through unsupervised learning \cite{fei2005bayesian,liu2011recognizing,huang2016unsupervised}. However, these learned concepts may not have true meaning in most cases. In this work, empowered by CLIP \cite{radford2021learning}, we design an unsupervised concept learning method that is able to 
learn a large set of visual concepts with true semantic meaning from images using a set of pre-defined text concepts.

\section{Method}
In this section, we present our method of cross-modal concept learning and inference (CCLI) in detail. 

\begin{figure*}[!th]
\centering
\includegraphics[width=\textwidth]{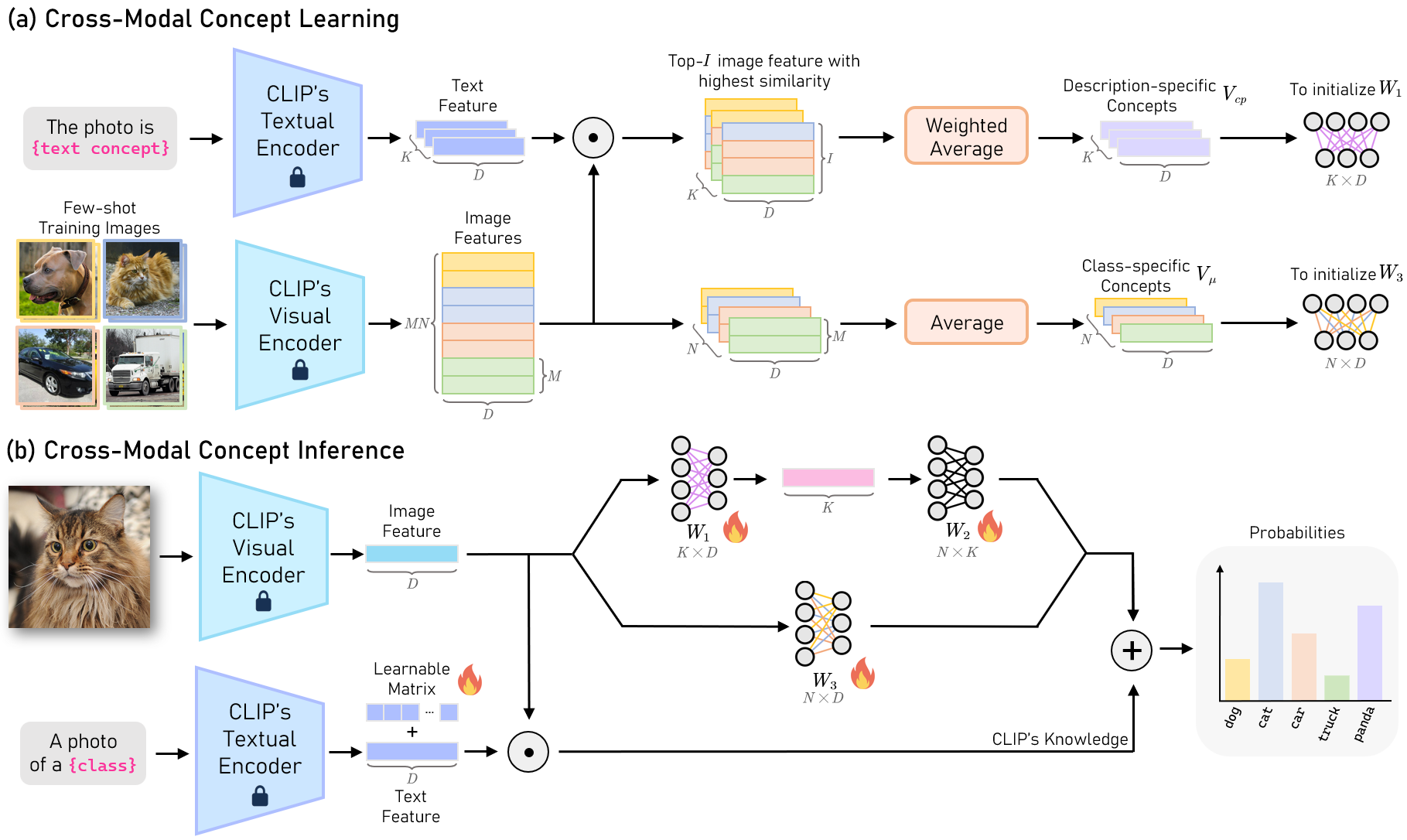}
\caption{\textbf{Overview of our Cross-modal Concept Learning and Inference (CCLI) method}. \textbf{(a)} shows the concept learning and \textbf{(b)} shows the concept inference process of our method.  The fire icon means the parameters will be updated during training.}
\label{fig:overview}
\vspace{-10pt}
\end{figure*}

\subsection{Revisiting the CLIP Model}
CLIP \cite{radford2021learning} consists of two parallel encoders, one for image and the other for text. It normally utilizes a ResNet \cite{he2016deep} or a ViT \cite{dosovitskiy2020image} as an image encoder, which maps an image into a visual  representation vector. The text encoder is a Transformer, which takes the text as input and generates a textual feature. During the training process, CLIP exploits a contrastive loss to enforce similarity between image-text pairs.
We denote CLIP's encoders as $\{E_t, E_v\}$, where $E_t$ is the text encoder and $E_v$ is the image encoder. After training, CLIP can be utilized for image classification in zero-shot scenarios with a hand-crafted prompt \cite{gao2021clip}. Given a  test image $X_{te}\in \mathbb{R} ^{C\times H\times W}$ of class $y$  for a $N$-class classification problem, in the zero-shot setting, we first append the class name text of every $y_i$ in $\{ y_i \}_{i=1}^N$ to a hand-crafted prompt denoted by $\pi$, such as $\pi =$``a photo of a", to build a class-specific text inputs $\{\pi; y_i\}$. Then, we generate the text features $\{ t_i \}_{i=1}^N$ using the text encoder $E_t$, where $t_i = E_t(\{\pi; y_i\})$. The cosine similarity score between the text feature $t_i$ and the image feature $v=E_v(X_{te})$ is given by
\begin{equation}
\label{eq-sim}
   \mathrm{sim}\left( t_i,v \right)=\frac{t_i \cdot v}{\Vert t_i \Vert  \Vert v \Vert}.
\end{equation}
The prediction probability on $X_{te}$ is computed as
\begin{equation}
\label{eq-clip}
   p(y = i|X_{te})=\frac{\exp \left( \mathrm{sim}\left( t_i,v \right) /\tau \right)}{\sum\nolimits_{j=1}^K{\exp \left( \mathrm{sim}\left( t_j,v \right) /\tau \right)}}, 
\end{equation}
where $\tau$ is the temperature hyper-parameter of the softmax function learned by CLIP.

\begin{figure}[!th]
\centering
\includegraphics[width=\linewidth]{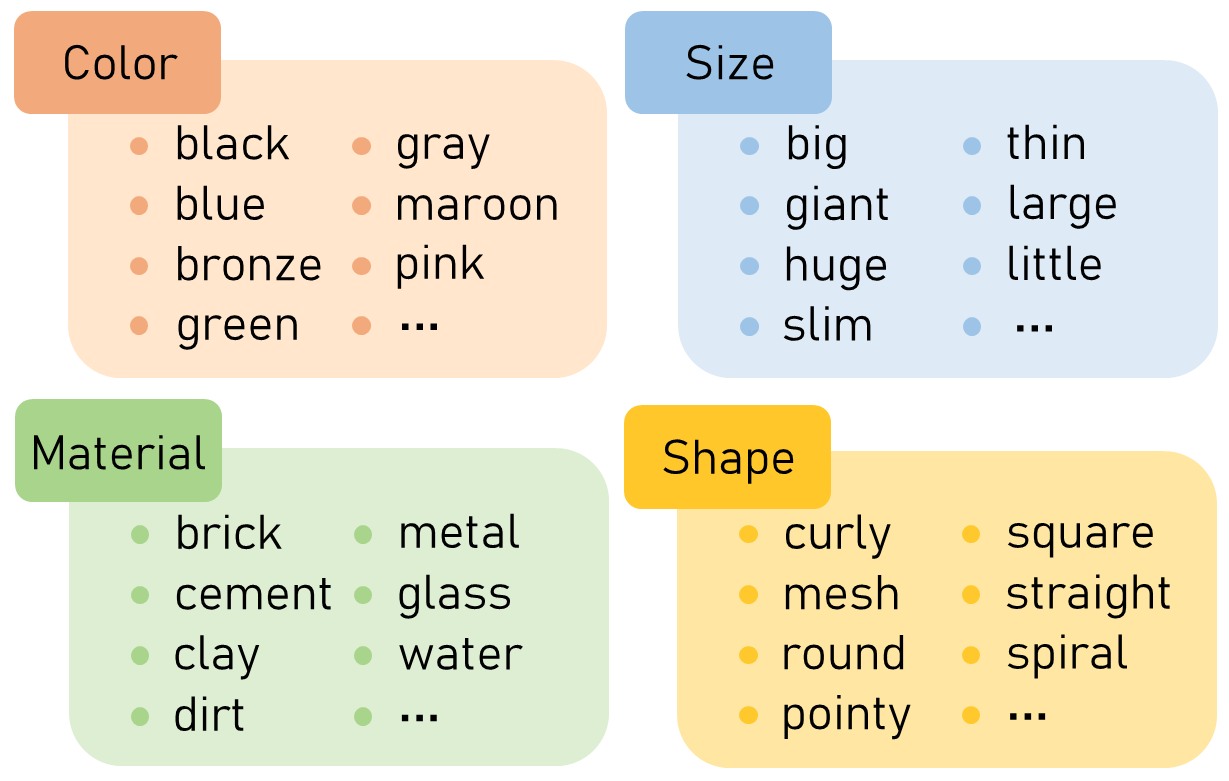}
\caption{\textbf{Dictionary of text concepts}. Here, we present some example words describing color, size, material and shape  in our dictionary.}
\label{fig:dictionary}
\vspace{-10pt}
\end{figure}

\subsection{Cross-Modal Concept Learning}
\label{sec:ADS}
As shown in Figure \ref{fig:overview}(a),  we first construct a comprehensive dictionary $\Omega_t$ of text concepts of size $K$ which describe major visual concepts in all images.
This dictionary contains $K=1000$ common text descriptions for visual attributes collected from existing visual attribute datasets \cite{zhao2019large,pham2021learning}, including words describing colors, textures, shapes, actions, materials, expressions, \textit{etc}. The set of text concepts is empirically designed, trying to include different types of descriptions of objects. Some example words in this dictionary are presented in Figure \ref{fig:dictionary}.  We denote this dictionary by $\Omega_t \triangleq \{d_i\}_{i=1}^K$. Following the zero-shot setting of CLIP, we first append $d_i$ to a hand-crafted prompt $\pi =$ ``The photo is" to build a concept-specific text input $\{\pi; d_i\}$. Then, we can generate text concept features $T \triangleq \{ t_i \}_{i=1}^K$ using the text encoder $E_t$, 
\begin{equation}
\label{eq-textf}
    t_i = E_t(\{\pi; d_i\}).
\end{equation}

In our proposed method for few-shot learning and domain generalization, the set of visual concepts is learned from the training images using the text concept features $T$ and the CLIP model. 
For example, for $M$-shot $N$-class few-shot learning, we have $M$ annotated images in each of the $N$ classes. The training set is denoted as $ X \triangleq\{x_j\}_{j=1}^{MN}$. Using the CLIP visual encoder $E_v$, 
we can generate their image features $V \triangleq \{ v_j \}_{j=1}^{MN}$, where $v_j = E_v(x_j)$. 

For every text concept feature $t$ in $T$, we calculate the similarity score between $t$ and every visual feature in $V$ by Equation (\ref{eq-sim}), $S_t = \mathrm{sim}\left( t,v_j \right)= tv_j$, in which both $t$ and $v_j$ are normalized. Thus, for each text concept feature  $t$, we have $MN$ similarity scores. Then, we select the top $I$ image features with the highest similarity scores. 
We compute the weighted average of these top $I$ image features as 
 \begin{equation}
\label{eq-weighted}
    \bar{v}=\frac{\sum\nolimits_{i=1}^I{w_iv_i}}{\sum\nolimits_{i=1}^I{w_i}},
\end{equation}
where $w_i$ is the CLIP similarity score between image feature $v_i$ and the text feature $t$.
In this way, for all text concepts, we have obtained their corresponding visual concepts, which can be denoted as 
 \begin{equation}
\label{eq-Vcp}
    V_{cp} \triangleq \{\bar{v_i}\}_{i=1}^K.
\end{equation}

\begin{figure}[!th]
\centering
\includegraphics[width=\linewidth]{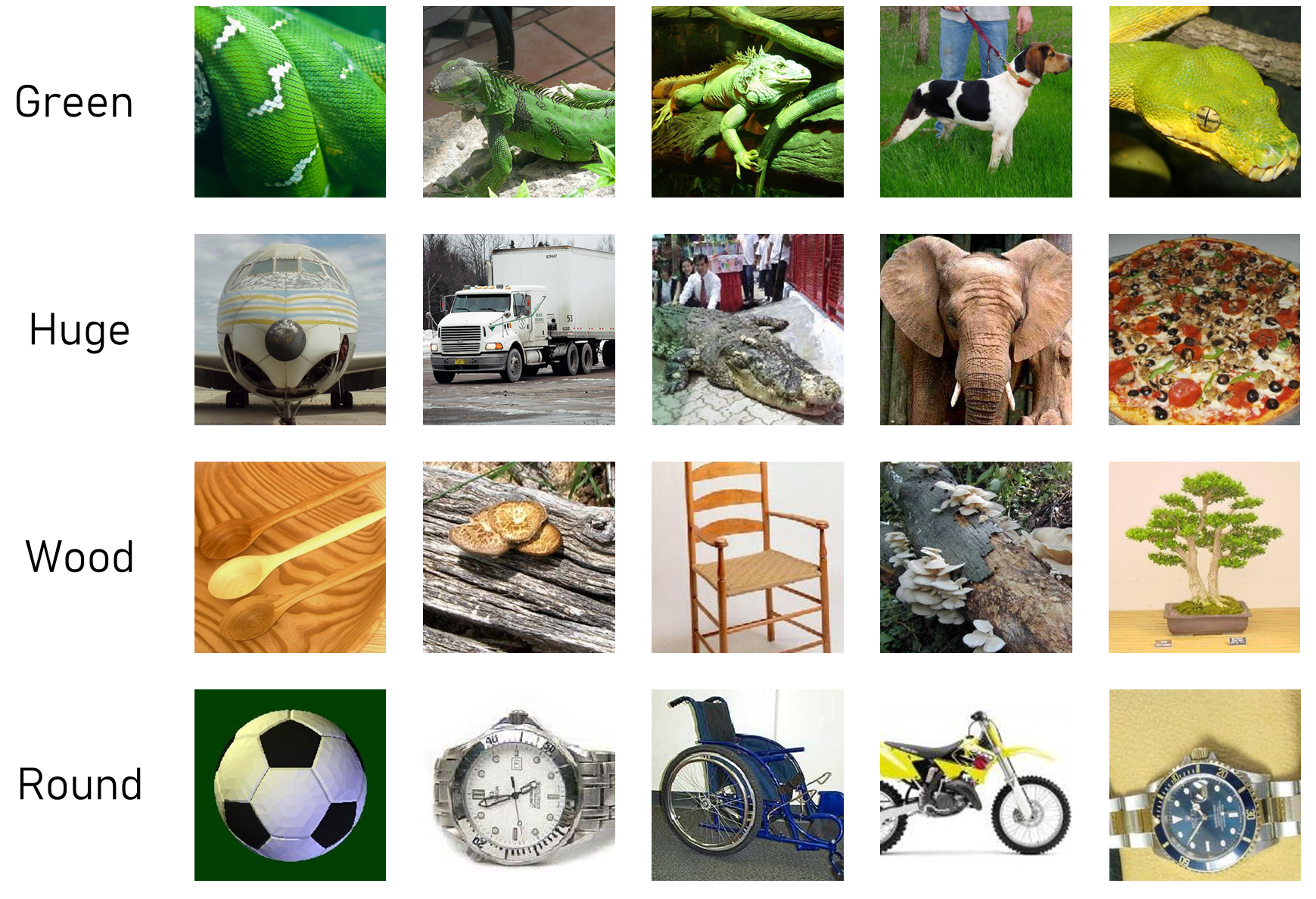}
\caption{\textbf{Examples of top 5 images in the concept learning process}. Here, we present images that are most related to the four concepts: green, huge, wood, and round.}
\label{fig:top}
\vspace{-10pt}
\end{figure}

In this work, this set of visual concepts is referred to as the description-specific visual concepts.
Figure \ref{fig:top} shows the top five images for four distinct text concepts selected by the concept learning process to show the effectiveness of our method.

 Besides the description-specific visual concepts, we also construct class-specific visual concepts. 
 Specifically, for each class of training images, we calculate the mean feature of $M$-shot images generated by the visual encoder. We then obtain $N$ class-specific features, 
 \begin{equation}
\label{eq-mu}
    V_{\mu} \triangleq \{ \mu_n \}_{n=1}^N.
\end{equation}

\subsection{Cross-Modal Concept Inference}

Once the collection of visual concepts is learned, during concept inference, we represent the input image using this set of visual concepts. As shown in Figure \ref{fig:overview}(b), based on this visual concept representation, we learn an inference network to classify the image. Our concept inference network consists of two parallel networks appended to the image encoder of CLIP. 

The first one is a two-layer network after the image encoder of CLIP. We initialize the first layer with weights $W_1 \in \mathbb{R}^{ K \times D}$ with $V_{cp}$, so that a higher concept score can be obtained when the input feature is consistent with a more compatible description-specific concept feature. Then, the second layer of the network (with weights of $W_2 \in \mathbb{R}^{N \times K}$) integrates all concept scores of the input image and performs effective concept inference. This two-layer network, which is a part of the concept inference model, can be denoted as
\begin{equation}
\label{eq-A}
    A\left( x \right) =W_2\left(\mathrm{ReLU}\left(W_1x\right)\right).
\end{equation}
During training, the weights $W_1$ and $W_2$ are updated by gradient descent. After supervised learning, the concept features can be optimized for a specific dataset to learn more discriminative concept-level representations. 
On top of the concept inference model, the affinities \cite{ru2022learning,zhang2022tip} can be further computed as
\begin{equation}
\label{eq-Aff}
    \mathrm{Aff}(x) =\exp \left( -\delta \left(1-x)\right) \right),
\end{equation}
where $\delta$ is a hyper-parameter for adjusting the sharpness, which controls the influence of the most compatible attribute-specific visual features on the final prediction. 
The exponential function is utilized to convert the outputs into non-negative values.
 Given the $L2$ normalized feature $v\in \mathbb{R}^{1 \times D}$ of the training image, which is generated by the visual encoder $E_v$, the logits $L_a \in \mathbb{R}^{1 \times N}$ of the concept inference model can be denoted as
\begin{equation}
\label{eq-La}
    L_a = \mathrm{Aff}\left(A\left( v \right)\right) =\exp \left(-\delta (1-\mathrm{ReLU}(vW_1^\top)W_2^\top) \right)
\end{equation}
and used for final category classification.

Similarly, the second network is simply a one-layer network used to provide class-specific concept inference, denoted as
\begin{equation}
\label{eq-q}
    Q(x) =W_3 \left( x \right) ,
\end{equation}
we initialize $W_3$ as class-specific visual concept $V_\mu$ from Equation (\ref{eq-mu}). According to Equation (\ref{eq-Aff}), the logits of class-specific concept inference can be denoted as 
\begin{equation}
\label{eq-Lq}
    L_q  =\exp \left(-\eta (1-vW_3^\top) \right)
\end{equation}
where $\eta$ is a hyper-parameter similar to $\delta$ for adjusting the sharpness.

\subsection{CCLI for Few-Shot Learning}
\label{sec:CCLI}
Inspired by CoOp and TaskRes \cite{zhou2022learning, yu2022task}, as illustrated in Figure \ref{fig:overview}, we enhance the original CLIP by appending a learnable matrix to the text features $f_t \in \mathbb{R}^{N \times D}$ generated by text encoder $E_t$. Unlike existing prompt learning methods, our method directly operates on the text features generated by the  text encoder, so there is no need to encode the text every time during training. This preserves the original knowledge of CLIP while also allowing for the acquisition of few-shot learning knowledge in an efficient manner. We define the text adapter as $\hat{f}_t=f_t+\beta Z$, where $Z$ is a learnable matrix with the same shape of $f_t$, $\beta$ is a hyper-parameter that controls how much of $Z$ we use to combine with $f_t$. The logit of enhanced CLIP is:
\begin{equation}
\label{eq-Le}
    L_e=v{\hat{f}_t}^\top=v(f_t+\beta Z)^\top.
\end{equation}
where $v$ is the image features generated by $E_v$. During training, $Z$ is updated by gradient descent. For each task, we learn a task-specific text adapter $Z$. In this way, we can preserve the prior knowledge of CLIP and obtain the knowledge from new tasks, so that CLIP can be better adapted to downstream tasks.

During few-shot learning, we combine the output logits of the concept inference and the text adapter, and the total logits of the input image $v$ used for the final classification are calculated as 
\begin{equation}
\begin{aligned}
\label{eq-LOGIT}
     \mathrm{Logits} &\ = \alpha L_a+ \lambda L_q+L_e  \\ &\
     =\alpha \exp \left(-\delta (1-\mathrm{ReLU}(vW_1^\top)W_2^\top) \right) \\ &\ + \lambda \exp \left(-\eta (1-vW_3^\top) \right)  +v(f_t+\beta Z)^\top.
\end{aligned}
\end{equation}
where $\alpha$ is a hyper-parameter that controls the ratio of different logits  from concept inference with enhanced CLIP. $\{W_1,W_2,Z\}$ represent all learnable parameters. The sensitivity levels of the hyper-parameters are evaluated in Section \ref{sec:ablation}. The pseudo-code of the proposed CCLI method is shown in Algorithm \ref{alg-method}.

\begin{table*}[htbp]
\centering
    \caption{\textbf{The detailed statistics of datasets used in experiments}. The first 11 datasets are used for few-shot learning evaluation, and the last 4 datasets are for domain generalization.}
    \label{tab:dataset}
    \resizebox{0.87\textwidth}{!}{
    \begin{tabular}{lcccc}
    \toprule
Dataset                   &  Classes   &  Training size   &  Testing size  &  Task \\ \midrule
Caltech101~\cite{fei2004learning}  &  100  &  4,128  &  2,465 &  Object recognition \\
DTD~\cite{cimpoi2014describing} &  47  &  2,820  &  1,692  &   Texture recognition\\ 
EuroSAT~\cite{helber2019eurosat} &  10  &  13,500  &  8,100  &  Satellite image recognition \\ FGVCAircraft~\cite{maji2013fine}  &  100  &  3,334  &  3,333  &  Fine-grained aircraft recognition\\
Flowers102~\cite{nilsback2008automated}  &  102  &  4,093  &  2,463  &  Fine-grained flowers recognition \\ Food101~\cite{bossard2014food}  &  101  &  50,500 &  30,300  &  Fine-grained food recognition  \\ ImageNet~\cite{recht2019imagenet}  &  1,000  &  1.28M  &  50,000  &  Object recognition \\ OxfordPets~\cite{parkhi2012cats}  &  37   &  2,944  &  3,669  &  Fine-grained pets recognition \\ StanfordCars~\cite{krause20133d}  &  196  &  6,509  &  8,041  &  Fine-grained car recognition \\
SUN397~\cite{xiao2010sun} &  397 &  15,880  &  19,850  &  Scene recognition\\ 
UCF101~\cite{soomro2012ucf101} &  101  &  7,639  &  3,783  &  Action recognition\\
\midrule
ImageNet-V2~\cite{recht2019imagenet}  &  1,000  &  -  &  10,000  &  Robustness of collocation  \\
ImageNet-Sketch~\cite{wang2019learning}  &  1,000  &  -  & 50,889  &  Robustness of sketch domain\\
ImageNet-A~\cite{hendrycks2021natural} &  200  &  -  & 7,500  & Robustness of adversarial attack\\
ImageNet-R~\cite{hendrycks2021many} &  200  &  -  & 30,000 & Robustness of multi-domains\\
    \bottomrule
    \end{tabular}
    }
\end{table*}

\begin{algorithm}[h]
\SetAlCapFnt{\small}
\SetAlCapNameFnt{\small}
\caption{Pseudocode of our CCLI method.}
\label{alg-method}
\LinesNumbered
\begin{small}
\KwIn{Pre-trained CLIP image and text encoder $E_v,E_t$\;}
\KwIn{The dictionary of text concepts $\Omega_t$\;}
\KwIn{Training set $\mathcal{D}^{tr}$ of target task\;}
\KwIn{Class names $\{c_i\}_{i=1}^N$ and hand-crafted prompt $\pi$\;}
Generate text concept features $T$ for all text concepts in dictionary $\Omega_t$ by Equation (\ref{eq-textf})\;
Learn description-specific concepts $V_{cp}$ by Equation (\ref{eq-Vcp})\;
Learn class-specific concepts $V_{\mu}$ by Equation (\ref{eq-mu})\;

Initialize $W_1$ with description-specific concepts $V_{cp}$\;
Initialize $W_3$ with class-specific concepts $V_\mu$\;
Initialize the task-specific text adapter $Z$ with zeros\;
\For{\_ \textbf{in} iterations}{
Sample a batch $\{(x_j,y_j)\}_{j=0}^J$ from $\mathcal{D}^{tr}$\;
Compute $f_t=\{E_t(\{\pi, c_i\})\}, i=1,\dots,N$\;
Compute $\hat{f}_t$ using the learnable matrix $Z$\; 
Compute $v=\{E_v(x_j)\}, j=1,\dots,J$\;
Let $Labels = \{ y_j \}_{j=1}^J$\;
Compute $L_a$, $L_q$ and $L_e$ according to Equation (\ref{eq-La}), (\ref{eq-Lq}) and (\ref{eq-Le}) \;
Compute $Logits$ according to Equation (\ref{eq-LOGIT})\;
Compute $loss=CrossEntropyLoss(Logits, Labels)$\;
Update $W_1$, $W_2$, $W_3$, $Z$ by gradient descent\;
}
\end{small}
\end{algorithm}




\section{Experimental Results}
In this section, we present performance comparisons with state-of-the-art methods on the few-shot learning and domain generalization tasks, and ablation studies to demonstrate the effectiveness of our proposed method. We summarize the detailed statistics of datasets used in experiments in Table \ref{tab:dataset}.

\subsection{Few-Shot Learning} \label{sec:fewshot}
The objective of few-shot learning is to transfer a trained model to novel tasks with limited available supervision. Pre-trained VLMs, such as CLIP, provide a new paradigm for this task.

\begin{table}[t]
\centering
\caption{\textbf{Few-shot classification accuracy (\%) on ImageNet~\cite{deng2009imagenet} of different methods with quantitative values}. The results marked as \textbf{bold} represent the highest performance, while the second-best results are indicated by being \underline{underlined}.}
\label{tb:imagenet}
\resizebox{\linewidth}{!}{
	\begin{tabular}{lcccccc}
	\toprule
		Few-shot Setup & 1    & 2  & 4  & 8 & 16 \\ \midrule
        Zero-shot CLIP~\cite{radford2021learning}  &60.33 &60.33 &60.33 &60.33 &60.33\\
	    Linear-probe CLIP~\cite{radford2021learning}  &22.17 &31.90 &41.20 &49.52 &56.13\\
		CoOp~\cite{zhou2022learning}  &57.21& 55.93& 59.88& 60.91& 62.26 \\
        CoCoOp~\cite{zhou2022conditional}  &60.78& \underline{61.91}& 62.49& 62.38& 62.70 \\
	    CLIP-Adapter~\cite{gao2021clip}  &61.20  & 61.52 &  61.84 &  62.68 &  63.59 \\
	    Tip-Adapter-F~\cite{zhang2022tip}  &\underline{61.32}   &61.69   &\underline{62.52} &\underline{64.00}   &\underline{65.51}  \\
        PLOT~\cite{chen2023plot} &59.54& 60.64& 61.49& 61.92& 63.01 \\
        DeFo~\cite{wang2023learning} &59.44 & 59.72 & 60.28 & 61.73 & 64.00\\ 
        \rowcolor{gray!20}
        \textbf{CCLI (Ours)} & \textbf{62.27}& \textbf{62.96}& \textbf{63.76}& \textbf{64.95}& \textbf{66.53}\\
	\bottomrule
	\end{tabular}
}
\end{table}

\begin{figure*}[ht]
\centering
\includegraphics[width=\textwidth]{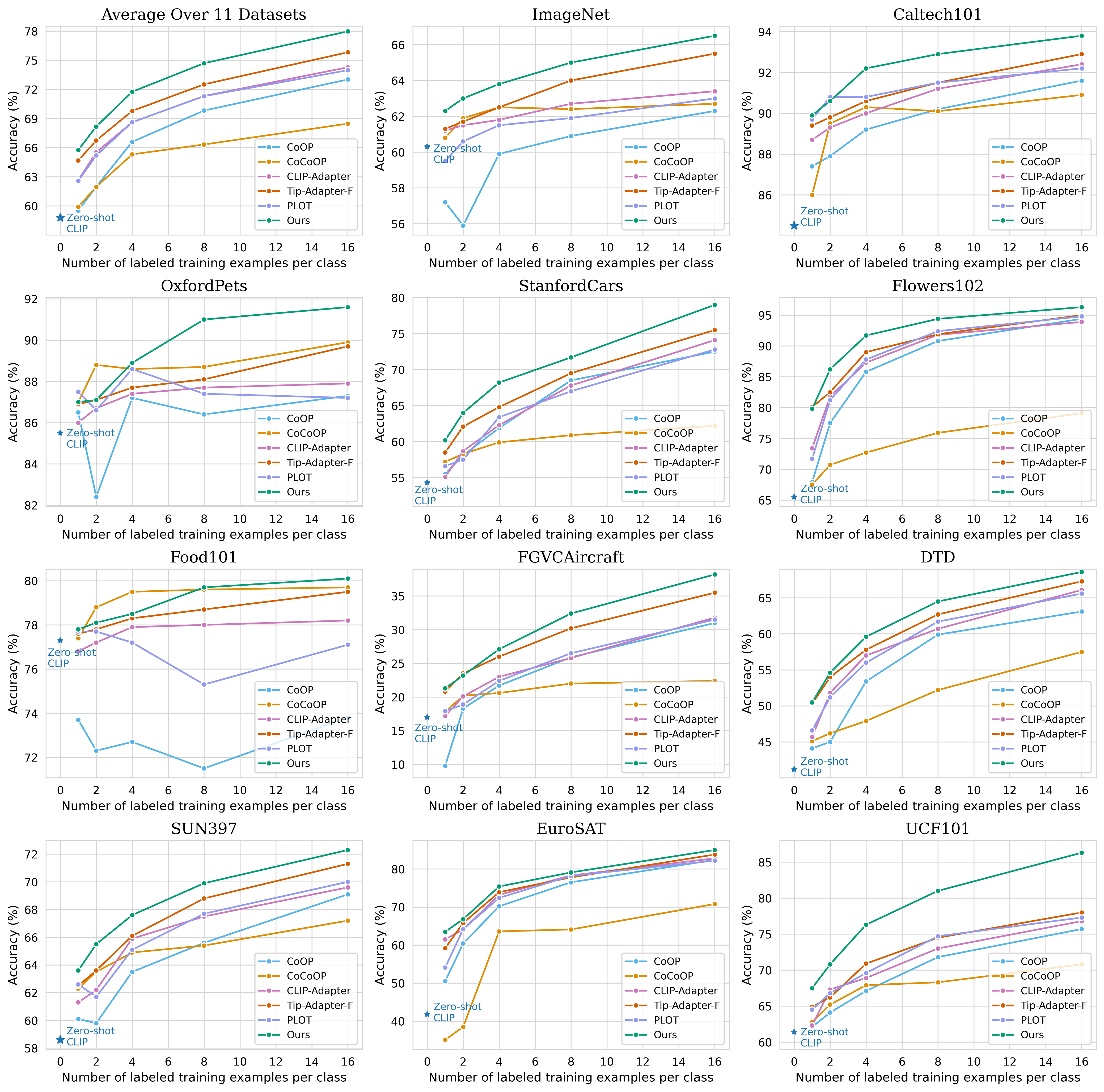}
\caption{\textbf{Classification performance comparison on few-shot learning}, {\em i.e.}, 1-/2-/4-/8-/16-shot, on 11 benchmark datasets. The top-left is the averaged accuracy over the 11 datasets.}
\label{fig:fewshot_results}
\end{figure*}

\subsubsection{Datasets} 
Following prior methods~\cite{zhou2022learning,zhang2022tip}, we adopt the few-shot evaluation protocol to assess our method on 11 widely-used image classification datasets in Table \ref{tab:dataset}, spanning the breadth of generic object classification (ImageNet~\cite{recht2019imagenet}, Caltech101~\cite{fei2004learning}), fine-grained object classification (OxfordPets~\cite{parkhi2012cats}, StandfordCars~\cite{krause20133d}, Flowers102~\cite{nilsback2008automated}, Food-101~\cite{bossard2014food}, FGCV Aircraft~\cite{maji2013fine}), texture classification (DTD~\cite{cimpoi2014describing}), remote sensing recognition (EuroSAT~\cite{helber2019eurosat}), scene recognition (SUN397~\cite{xiao2010sun}), and action recognition (UCF101~\cite{soomro2012ucf101}). These datasets provide a comprehensive benchmark to evaluate the few-shot learning performance for each method.

\subsubsection{Comparison Methods}
We compare our method with eight baseline methods reviewed in Section \ref{sec:related}: zero-shot CLIP~\cite{radford2021learning}, linear probe CLIP~\cite{radford2021learning}, CoOp~\cite{zhou2022learning}, CoCoOp~\cite{zhou2022conditional}, CLIP-Adapter~\cite{gao2021clip}, Tip-Adapter-F~\cite{zhang2022tip}, PLOT~\cite{chen2023plot} and DeFo~\cite{wang2023learning}. Therein, zero-shot CLIP relies on manually designed prompts. For a fair comparison, we choose CoOp's best-performance setting - with the class token placed at the end of 16-token prompts. We also choose the best variant of CLIP-Adapter and the fine-tuned version of Tip-Adapter (Tip-Adapter-F) in our experiments.

\subsubsection{Implementation Details}
Our model is built upon the publicly available CLIP model. We use the ResNet-50 image encoder and transformer text encoder as the CLIP backbone. Throughout the training process, we keep both the visual and text encoders frozen. We follow the data pre-processing protocol in CLIP, including operations of resizing, random cropping, \textit{etc}.  In our experiments, the hyper-parameter $I$ to control the number of top visual features is set to 5. We train our model for 100 epochs on ImageNet and 50 epochs on other datasets. The text feature adapter, which is a learnable matrix with the same shape as the text features generated by the text encoder, is initialized with zeros. We set $\beta$ in Equation (\ref{eq-Le}) to $0.8$ for ImageNet and $0.6$ for the rest datasets.  We use a batch size of 256 and an initial learning rate of $10^{-3}$. Our models are optimized by AdamW~\cite{kingma2014adam} optimizer with a cosine annealing scheduler. One single NVIDIA RTX 3090 GPU is used for training. 
We adhere to the conventional evaluation protocol for few-shot learning, where training involves a random selection of 1, 2, 4, 8, and 16 shots per class, followed by testing on the complete test set.

\subsubsection{Performance Results}
In Table \ref{tb:imagenet}, we compare the few-shot classification accuracy of our method on ImageNet~\cite{deng2009imagenet} with other state-of-the-art methods. Our proposed method obtains promising results in this dataset and an average of $+1\%$ improvement can be observed in all few-shot settings.

Figure \ref{fig:fewshot_results} shows the comparison with five baseline methods on all 11 datasets, and the average accuracy is shown in the top-left sub-figure of Figure \ref{fig:fewshot_results}. We observe that our method performs the best in few-shot learning and obtains the highest test accuracy on average over other state-of-the-art methods. Notably, with the increase in the number of shots, the performance gain over other methods becomes larger. This proves that training with more shots enables our model to build a more robust and discriminative concept-level representation. In comparison to zero-shot CLIP, our method consistently surpasses it by huge margins on all datasets. In Figure \ref{fig:fewshot_results}, our method performs worse on 1 and 2 shots for OxfordPets and Food101 datasets. This is because, when computing the class-specific concepts, we only have 1 or 2 images to represent the whole class. This is not fully effective and degrades the performance. However, the overall performance has shown the effectiveness of our proposed CCLI method.

\begin{figure*}[ht]
\centering
\includegraphics[width=\textwidth]{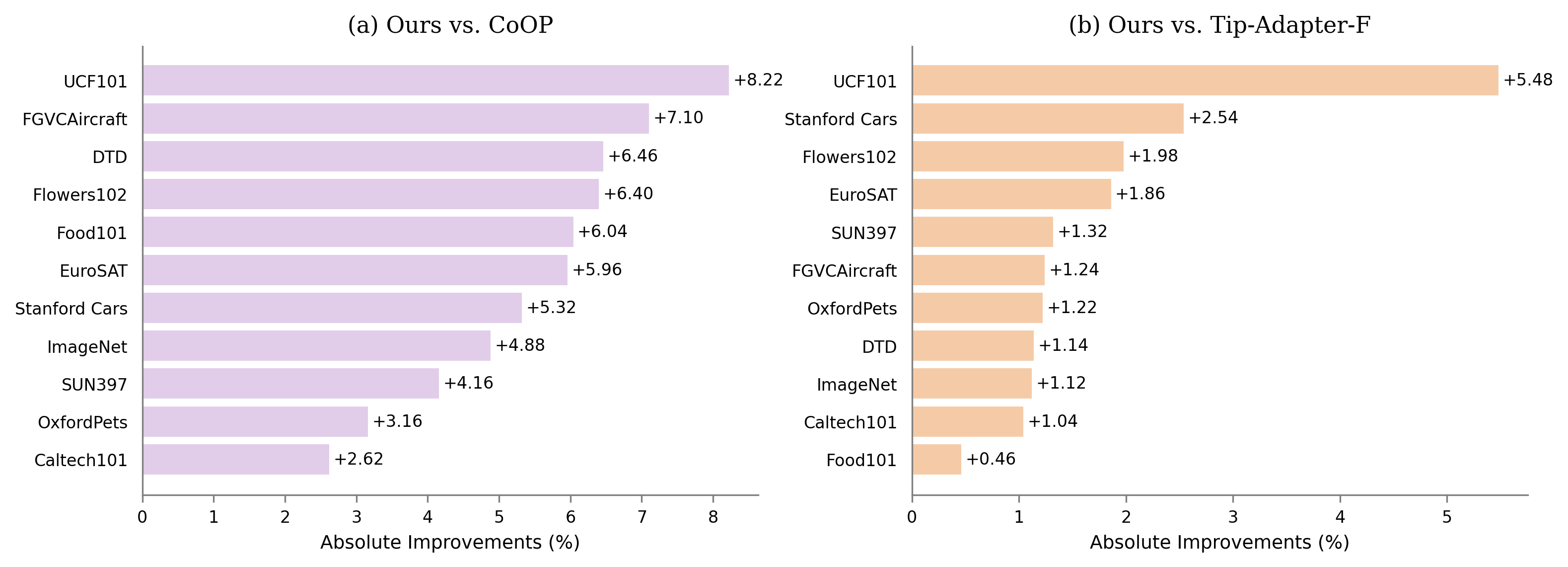}
\caption{\textbf{Comparison with CoOp~\cite{zhou2022learning} and Tip-Adapter-F~\cite{zhang2022tip}}.We show the absolute improvement of our method compared with prompt tuning method (CoOp) and adapter-based method (Tip-Adapter-F). These comparisons are conducted by their average results of few shots (1, 2, 4, 8, and 16) per category.}
\label{fig:improvement}
\end{figure*}

\textbf{Comparison with prompt tuning methods.}  
Compared to CoOp~\cite{zhou2022learning}, which is one of the state-of-the-art prompt learning methods, our approach consistently yields better recognition performance. The absolute performance improvement compared to CoOp on each dataset is shown on the left side of Figure \ref{fig:improvement}. We can see that the largest gain over CoOp is $+8.2\%$ on UCF101, and even the smallest gain is $+2.6\%$. Furthermore, our method outperforms CoCoOp by a huge margin on the average performance as indicated in Figure \ref{fig:fewshot_results}. This demonstrates that our methods yield superior performance against the prompt learning methods.

\textbf{Comparison with adapter-style methods.} 
As shown in the top-left sub-figure of Figure \ref{fig:fewshot_results}, our method exhibits significantly superior performance compared to the CLIP-Adapter~\cite{gao2021clip} and Tip-Adapter~\cite{zhang2022tip} on these  11 datasets. Compared to the CLIP-Adapter, our method obtains superior performance on all the datasets. Tip-Adapter-F is the top-performing method with adapter style. Our method outperforms Tip-Adapter-F by an average of $+1.8\%$  on all datasets. The largest gain over Tip-Adapter-F is $+8.3\%$ on the UCF101 with a 16-shot setting. The absolute performance improvements compared to Tip-Adapter-F on each dataset are shown on the right side in Figure \ref{fig:improvement}. We can see that our model achieves the largest performance gain of $+5.5\%$ over Tip-Adapter-F on UCF101. 
Overall, our method substantially outperforms its baselines in few-shot learning tasks. These findings serve as strong evidence showcasing the effectiveness of our approach.

\subsection{Domain Generalization}
Robustness to distribution shift is critical for the generalization ability of machine learning models. Pre-trained VLMs such as CLIP have exhibited strong robustness to distribution shifts.

\subsubsection{Experimental Settings}
We evaluate the domain generalization performance of our method by 16-shot training on ImageNet~\cite{deng2009imagenet} and testing on four ImageNet variant datasets: ImageNet-V2~\cite{recht2019imagenet}, ImageNet-Sketch~\cite{wang2019learning}, ImageNet-A~\cite{hendrycks2021natural}, and ImageNet-R~\cite{hendrycks2021many} in Table \ref{tab:dataset}. ImageNet-V2~\cite{recht2019imagenet} serves as a replicated test set comprising 10,000 natural images obtained from an alternative source. This collection encompasses 1,000 ImageNet classes. ImageNet-Sketch~\cite{wang2019learning} encompasses a dataset of 50,000 monochrome sketch images, all of which belong to the same set of 1,000 ImageNet classes.  
ImageNet-A~\cite{hendrycks2021natural} is a collection of naturally adversarially filtered images, featuring 7,500 pictures from 200 classes selected out of the original 1,000 classes found in ImageNet.
ImageNet-R~\cite{hendrycks2021many} is a dataset containing images with artistic renditions of ImageNet categories, including 30,000 images belonging to 200 of ImageNet's 1,000 categories.

\begin{tiny}
\begin{table*}[!ht]
\small
\begin{center}
\caption{\textbf{Comparison with other methods on robustness ($\%$) to natural distribution shifts (from ImageNet to ImageNet-V2/-Sketch/-A/-R)}. The results marked as \textbf{bold} represent the highest performance, while the second-best results are indicated by being \underline{underlined}.}
\label{table:generalization}
\resizebox{0.93\textwidth}{!}{
\begin{tabular}{lccccccc}
\toprule
\multirow{2}*{Method} & \multirow{2}*{Visual Backbone} & Source & \multicolumn{5}{c}{Target} \\ \cmidrule(lr){3-3} \cmidrule(lr){4-8} & & ImageNet & -V2 & -Sketch & -A & -R  & OOD Average \\
\midrule
Zero-Shot CLIP~\cite{radford2021learning} &\multirow{10}*{ResNet-50} &  60.33 &  53.27  & \underline{35.44}  & 21.65 &  56.00  & 41.59\\
Linear Probe CLIP~\cite{radford2021learning} & & 56.13  & 45.61  & 19.13  & 12.74  & 34.86 &  28.09\\
CoOp~\cite{zhou2022learning} &  & 63.33  & 55.40  & 34.67  & 23.06  & 56.60 &  42.43\\
CoCoOp~\cite{zhou2022conditional} &  & 62.81  & 55.72  & 34.48  & 23.32  & 57.74  & 42.82\\
ProGrad~\cite{zhu2022prompt} &  & 62.17  & 54.70  & 34.40  & 23.05  & 56.77  & 42.23\\
PLOT~\cite{chen2023plot} &  & 63.01  & 55.11  & 33.00  & 21.86  & 55.61  & 41.40\\
DeFo~\cite{wang2023learning} &   & \underline{64.00}  & \textbf{58.41}  & 33.18  & 21.68  & 55.84  & 42.28\\
TPT~\cite{shu2022tpt} &  & 60.74  & 54.70  & 35.09  & \underline{26.67}  & \underline{59.11}  & \underline{43.89}\\
\rowcolor{gray!20}
\textbf{CCLI (Ours)} &  & \textbf{66.53}  & \underline{58.18}  & \textbf{37.17}  & \textbf{30.93}  & \textbf{59.79}  & \textbf{46.52}\\
\midrule
Zero-Shot CLIP~\cite{radford2021learning}  &\multirow{8}*{ViT-B/16} &  67.83  & 60.83  & 46.15  & 47.77  & 73.96  & 57.18\\
Linear Probe CLIP~\cite{radford2021learning} & & 65.85  & 56.26 &  34.77 & 35.68 &  58.43  & 46.29\\
CoOp~\cite{zhou2022learning} &   & \underline{71.51}  & \underline{64.20}  & 47.99 &  49.71  & 75.21  & 59.28 \\
CoCoOp~\cite{zhou2022conditional} &   & 71.02  & 64.07  & \underline{48.75}  & 50.63  & 76.18  & 59.91\\
ProGrad~\cite{zhu2022prompt} &  & 70.45  & 63.35  & 48.17  & 49.45  & 75.21  & 59.05\\
TPT~\cite{shu2022tpt} &   & 68.98  & 63.45  & 47.94  & \underline{54.77}  & \underline{77.06}  & \underline{60.81}\\
\rowcolor{gray!20}
\textbf{CCLI (Ours)} &    & \textbf{74.57}  & \textbf{67.15}  & \textbf{49.78}  & \textbf{58.03}  & \textbf{77.83}  & \textbf{63.20}\\
\bottomrule
\end{tabular}
}
\end{center}
\end{table*}
\end{tiny}

\subsubsection{Comparison Methods}

We include nine previous methods reviewed in Section \ref{sec:related} for comparisons: zero-shot CLIP~\cite{radford2021learning}, linear probe CLIP~\cite{radford2021learning}, CoOp~\cite{zhou2022learning}, CoCoOp~\cite{zhou2022conditional}, ProGrad~\cite{zhu2022prompt}, PLOT~\cite{chen2023plot}, DeFo~\cite{wang2023learning}, TPT~\cite{shu2022tpt}, TPT + CoOp~\cite{shu2022tpt}. Therein, TPT + CoOp  is a method that applies TPT to prompts learned by CoOp and performs better than standalone TPT.

\subsubsection{Performance Results}
Table \ref{table:generalization} summarizes the results with two different visual backbones: ResNet-50 and ViT-B/16. We report the classification accuracy of the source domain (ImageNet), target domain (ImageNet-V2, ImageNet-Sketch, ImageNet-A, ImageNet-R), and the target average accuracy (OOD Average). We can see that  our method outperforms all other methods in most scenarios, which shows our model's remarkable robustness to distribution shifts. 

\subsection{Ablation Studies}
\label{sec:ablation}
To systematically evaluate our proposed method, we provide an empirical analysis of our design choices and illustrate the effects of different components of our method in this section. Ablations on the visual backbones and the number of shots are also reported in this section. All the experiments are conducted on ImageNet.

\textbf{Contributions of major algorithm components.}
Our method has two major new components, namely concept inference (CI) and text adapter (TA) in Section \ref{sec:CCLI}. As shown in Table \ref{table:adapters}, we find that both components contributes significantly to the overall performance. 

\begin{table}
\vspace{-20pt}
\caption{\textbf{Effectiveness of different components in our method}. CI is concept inference, and TA represents the text adapter.}
\label{table:adapters}
\centering
\resizebox{\linewidth}{!}{
\begin{tabular}{lccccc}
\toprule
Few-shot Setup & 1    & 2  & 4  & 8 & 16 \\ \midrule
Zero-shot CLIP                    & 60.33 & 60.33 & 60.33 & 60.33 & 60.33 \\
\, + CI               & 62.06 & 62.59 & 63.60 & 64.73 & 66.38 \\
\rowcolor{gray!20}
\, + CI + TA    & \textbf{62.27} & \textbf{62.96} & \textbf{63.76} & \textbf{64.95} & \textbf{66.53} \\
\bottomrule
\end{tabular}
}
\end{table}

\textbf{Description-specific and class-specific visual concepts.}
Table \ref{table:concept} shows the few-shot accuracy on ImageNet~\cite{deng2009imagenet} obtained by removing description-specific and class-specific visual concepts from our proposed model. The results show the significant contributions of both types of concepts to the overall performance. Removing either of these two components leads to a noticeable drop in accuracy, highlighting the importance of both description-specific and class-specific visual concepts in our method.

\begin{table}[htbp]
\vspace{-20pt}
\caption{\textbf{Effectiveness of description-specific and class-specific concepts in our method}. We report the accuracy without each kind of concept on ImageNet~\cite{deng2009imagenet} dataset. $V_{\mu}$ and $V_{cp}$ represent class-specific and description-specific concepts, respectively.}
\label{table:concept}
\centering
\resizebox{\linewidth}{!}{
\begin{tabular}{lccccc}
\toprule
Few-shot Setup & 1    & 2  & 4  & 8 & 16 \\ \midrule
Ours                  & 62.27 & 62.96 & 63.76 & 64.95 & 66.53 \\
\, w/o $V_{\mu}$  & 61.73 & 62.29 & 62.37 & 62.93 & 64.02 \\
\, w/o $V_{cp}$  & 61.22 & 61.48 & 62.11 & 62.28 & 63.35 \\
\bottomrule
\end{tabular}
}
\end{table}

\textbf{Visual backbones.}
Table \ref{table:backbones} summarizes the results on 16-shot ImageNet using various visual backbones containing ResNets and ViTs. It is observed that our approach demonstrates superior performance with more advanced visual backbones. In addition, no matter which visual backbone is used, our method shows consistently outperforms other methods. 

\begin{table}[ht]
\caption{\textbf{Evaluation of various visual backbones on ImageNet}. We report the results using a 16-shot setting for training.}
\label{table:backbones}
\centering
\resizebox{\linewidth}{!}{
\begin{tabular}{lcccc}
\toprule
Backbone& ResNet-50 & ResNet-101 & ViT-B/32 & ViT-B/16 \\
\midrule
Zero-shot CLIP         & 60.33     & 62.53      & 63.80    & 67.83    \\
CLIP-Adapter          & 63.59     & 65.39      & 66.19    & 71.13    \\
Tip-Adapter-F           & 65.51     & 68.56      & 68.65    & 73.69    \\
\rowcolor{gray!20}
\textbf{CCLI (Ours)}         & \textbf{66.53}     & \textbf{69.36}      & \textbf{69.60}    & \textbf{74.57}    \\
\bottomrule
\end{tabular}
}
\end{table}

\begin{figure}[t]
\centering
\includegraphics[width=\linewidth]{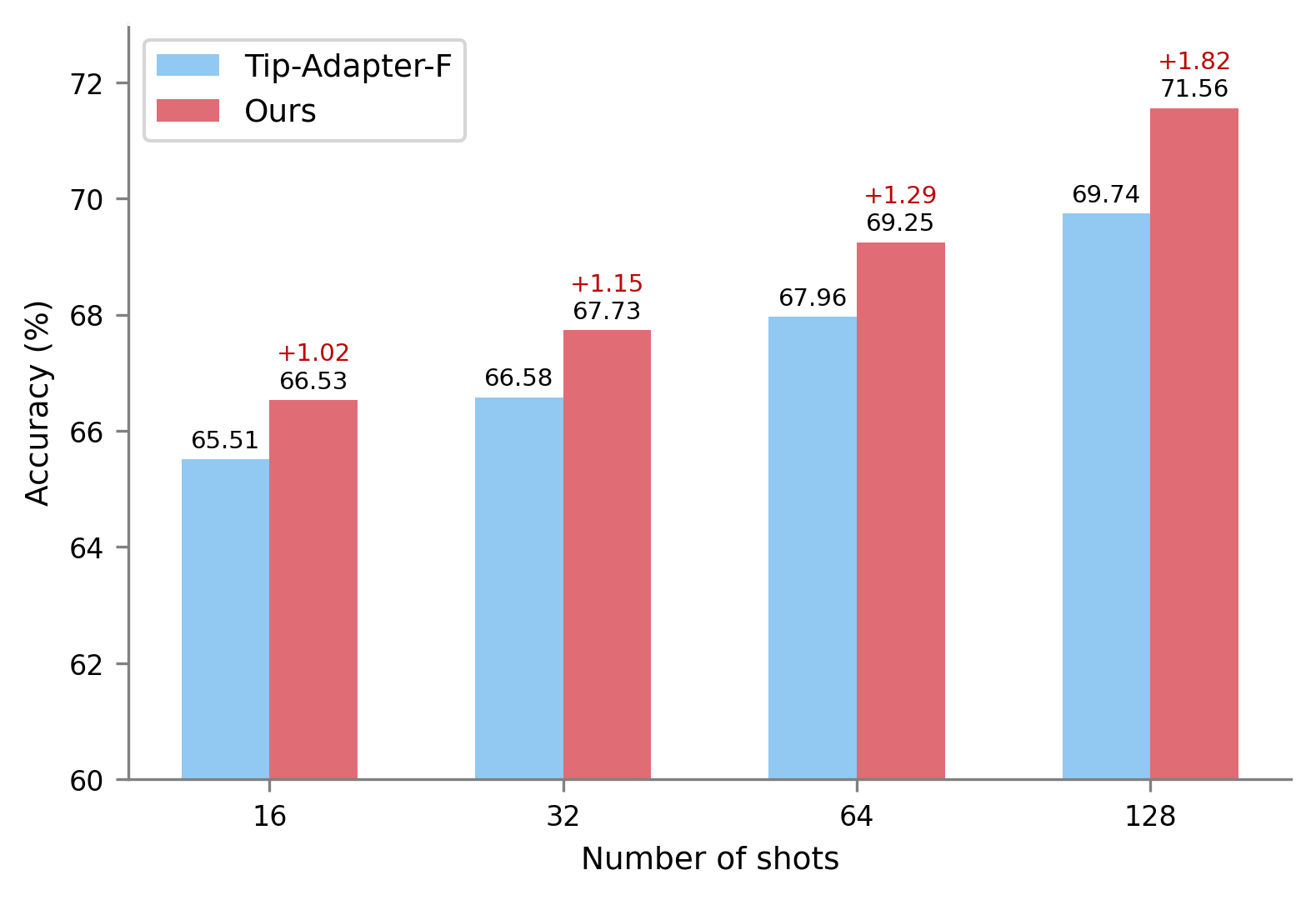}
\caption{\textbf{More shots for training.} We compare the performance of training with more shots (32, 64, 128). The experiments are conducted with ResNet-50 visual backbone.}
\label{fig:moreshot}
\end{figure}

\textbf{More shots for training.}
The result is shown in Figure \ref{fig:moreshot}. Our method achieves remarkable performance with more than 16 shots for training. As the number of shots grows, our method obtains more improvement in recognition accuracy. Compared to Tip-Adapter-F \cite{zhang2022tip}, our method achieves significant performance gains ranging from $+1.02\%$ (16-shot) to $+1.82\%$ (128-shot).  This suggests that, as the number of shots increases, our cross-modal concept learning and inference method becomes more robust and accurate.

\textbf{Sensitivity of hyper-parameters.}
In our experiments on ImageNet~\cite{deng2009imagenet}, we set the hyper-parameters $\alpha$, $\delta$, and $\beta$ to $1.5$, $4.5$, and $0.8$, respectively. To analyze the sensitivity of our model to these hyper-parameters, we conducted experiments by varying each of them and evaluated their impact on the model's performance. Table \ref{tb:hyper} shows that the value of $\alpha$, which controls the ratio of different components in the final logit, has a significant impact on the model's performance. When $\alpha$ is set to 0, the method degrades to zero-shot CLIP with only a text adapter. A moderate value of $1.5$ leads to the best performance for our model.
The hyper-parameter $\delta$, which controls the sharpness, has a relatively limited impact on performance. Our experiments show that the best performance is achieved when we set $\delta = 4.5$.
The sensitivity analysis of hyper-parameter $\beta$ indicates that its influence on the model's performance is minor. Varying $\beta$ had only a negligible effect on the model's performance.
Finally, we also conduct an ablation study of $I$ on the 16-shot ImageNet with $I = 1, 3, 5, 7, 9$. The accuracy varies from $63.78\%$ to $66.53\%$ and $I=5$ yields the optimal performance.

\begin{table}[H]
\caption{\textbf{Sensitivity of hyper-parameters}. All the results are reported on a 16-shot setting on ImageNet~\cite{deng2009imagenet}.}
\centering
\begin{adjustbox}{width=0.98\linewidth}
	\begin{tabular}{c|cccccc}
	\toprule
		\multicolumn{7}{c}{Sensitivity of Hyper-parameters} \\ 
		\midrule
		\multirow{2}{*}{$\alpha$}  &0.0 &0.5 &1.0 &\textbf{1.5} &2.0 &2.5 \\  
        \cmidrule(lr){2-7}
		 &61.60  & 64.21  &65.67 &\textbf{66.53} &64.92 &64.71\\ 
        \midrule
		
	    \multirow{2}{*}{$\delta$} &0.5 &2.5 &\textbf{4.5} &6.5 &8.5 &10.5 \\
	     \cmidrule(lr){2-7}
	    & 65.97  &66.26 &\textbf{66.53} &66.23 &66.11 &66.05\\
	    
	   \midrule
	   \multirow{2}{*}{$\beta$}&0.1 &0.2 &0.4 &0.6 &\textbf{0.8} &1.0 \\
	     \cmidrule(lr){2-7}
	    & 66.38  &66.42 &66.48 &66.50 &\textbf{66.53} &66.40\\
	   \midrule
	   \multirow{2}{*}{$I$}&1 &3 &\textbf{5} &7 &9 &11 \\
	     \cmidrule(lr){2-7}
	    & 63.78  &65.42 &\textbf{66.53} &66.37 & 66.18 &65.94\\
	    
	\bottomrule
	\end{tabular}
\end{adjustbox}
\label{tb:hyper}
\end{table}

\subsection{Complexity Analysis}
Table \ref{tab:efficiency} compares the performance and training time of our proposed method with state-of-the-art methods for 16-shot image classification on ImageNet~\cite{deng2009imagenet}. Based on the information provided in the table, it is evident that our approach significantly improves the accuracy while requiring relatively short training time.

\begin{table}[H]
\centering
\caption{\textbf{Efficiency and accuracy for different methods on ImageNet-16-shot.} The experimental evaluations are conducted using a batch size of 32 on a single NVIDIA GeForce RTX 3090 GPU. The last column reports the performance gain of each method over zero-shot CLIP.}
\renewcommand{\arraystretch}{1.1}
\resizebox{\linewidth}{!}{
\begin{tabular}{lccccccc}
\toprule 
Method  &  Epochs  &  Time  &  Accuracy  &  Gain \\
\midrule
Zero-shot CLIP~\cite{radford2021learning}  &  0  &  0  &  60.33  &  0 \\
Linear Probe CLIP~\cite{radford2021learning}  &   -  &  13min  &  56.13  &  -4.20 \\
CoOp~\cite{zhou2022learning}  &   200  &  14h 40min  &  62.26   &  +1.93 \\
ProGrad~\cite{zhu2022prompt}  &   200  &  17hr  &  63.45   &  +3.12 \\
CLIP-Adapter~\cite{gao2021clip}   &   200  &  50min  &  63.59   &  +3.26 \\
Tip-Adapter-F~\cite{zhang2022tip}   &   20  &  5min  &  65.51  &  +5.18 \\
\rowcolor{gray!20}
Ours  &   20  &  \textbf{4min}  &  \textbf{66.53}  &  \textbf{+6.20} \\
    \bottomrule
    \end{tabular}
}
\label{tab:efficiency}
\end{table}

\section{Conclusion}

The major contributions of this work can be succinctly outlined as follows. (1) We explore the powerful capabilities of CLIP in correlating texts and images and develop a new method to automatically learn  visual concepts from training images 
based on a collection of semantic text concepts. 
(2) Based on these  visual concepts, we are able to construct a discriminative representation of images and learn a concept inference network to perform downstream tasks.
(3) Extensive experimental results on few-shot image classification and domain generalization have demonstrated our proposed CCLI method outperforms the current state-of-the-art methods by large margins. 

The proposed idea can be naturally incorporated into other CLIP-based visual learning tasks, such as visual question answering, image captioning, and visual grounding. In the future, we hope to apply our approach to these tasks.

\printcredits

\bibliographystyle{cas-model2-names}

\bibliography{cas-refs}



\end{document}